\def\year{2015}
\documentclass[letterpaper]{article}
\usepackage{aaai}
\usepackage{times}
\usepackage{helvet}
\usepackage{courier}
\usepackage{natbib}
\begin{document}
\sloppy
\title{Hard to Cheat: A Turing Test based on Answering Questions about Images}
\author{Mateusz Malinowski and Mario Fritz \\
Max Planck Institute for Informatics \\
Saarbr{\"u}cken, Germany \\
\texttt{\{mmalinow,mfritz\}@mpi-inf.mpg.de} 
}
\maketitle
\begin{abstract}
\begin{quote}
Progress in language and image understanding by machines has sparkled the interest of the research community in more open-ended, holistic tasks, and refueled an old AI dream of building intelligent machines.
We discuss a few prominent challenges that characterize such holistic tasks and argue for ``question answering about images'' as a particular appealing instance of such a holistic task. In particular, we point out that it is a version of a Turing Test that is likely to be more robust to over-interpretations and contrast it with tasks like grounding and generation of descriptions.  Finally, we discuss tools to measure progress in this field.
 % by comparing different methods.
% As language and visual understanding by machines progresses rapidly,  we are observing an increasing
% interest in architectures that tightly interlink both modalities in a joint learning and inference process.
% This trend has allowed the community to progress towards more challenging and open tasks and refueled the hope at achieving the old AI dream of building machines that could pass a turing test in open domains.
% In order to steadily make progress towards this goal, we realize that quantifying performance becomes increasingly difficult. Therefore we ask how we can precisely define such challenges and
% how we can evaluate different algorithms on this open tasks?
% In this paper, we summarize and discuss such challenges as well as try to give answers where appropriate options are available in the literature.
% We exemplify some of the solutions on a recently presented dataset of  question-answering task based on real-world indoor images that establishes a visual turing challenge.
% Finally, we argue despite the success of unique ground-truth annotation, we likely have to step away from carefully curated dataset and rather rely on 'social consensus' as the main driving force to create suitable benchmarks. Providing coverage in this inherently ambiguous output space is an emerging challenge that we face in order to make quantifiable progress in this area.
\end{quote}
\end{abstract}
\section{Introduction}
\label{section:introduction}
Progress in machine perception and language understanding (e.g. \citep{krizhevsky2012imagenet, liang2013learning}) 
has inspired researchers to work on holistic tasks that interlink both modalities together
in a complex chain of perception, representation and inference.
% comment: representation means all the information extracted from language and vision together
 % and require the whole chain of perception, language representation, to deduction.
 Examples include: grounding 
% \citep{matuszek2012joint}
\citep{krishnamurthy2013jointly}, 
language generation \citep{karpathy2014generate, donahue2014long}, retrieval \citep{karpathy2014deep, malinowski14poolingretrieval}, 
and question answering about images \citep{malinowski14nips, malinowski14visualturing}.

% Recently we witness a tremendous progress in the machine perception \cite{krizhevsky2012imagenet, gupta2014learning, girshick2014rcnn, pishchulin2013strong,tompson2014joint, he2014spatial,lee2014deeply,simonyan2014very} and in the language understanding \cite{BlackburnBos:2005,zettlemoyer2007online, kwiatkowski2010inducing, mikolov2013distributed,cho2014learning} tasks.
% The progress in both fields has inspired researchers to build holistic architectures for challenging grounding  \cite{matuszek2012joint, krishnamurthy2013jointly}, natural language generation from image/video \cite{farhadi2010every,kulkarni2011baby,senina2014coherent}, image-to-sentence alignment \cite{socher2013grounded,karpathy2014deep,mao2014explain,kong2014you}, and recently presented question-answering problems \cite{liang2013learning,berant2014semantic,iyyer2014neural,faderopen,malinowski14nips}.

% Among many attempts to deviate
% \citep{shan2013visual, lake2013one}
% \citep{battaglia2013simulation} from the famous Turing Test \citep{turing1950computing},
Recently, \cite{malinowski14nips} have presented an approach for question answering about images that resembles the famous Turing Test \citep{turing1950computing}, while \cite{malinowski14visualturing} further discuss some of the associated challenges and issues. In the following, we 
 % behind a Visual Turing Challenge
 elaborate on data acquisition, contrast this challenge with other tasks including grounding, language generation, as well as highlight properties like robustness to over-interpretation, which makes it hard to cheat such a test.

\section{Challenges}
Architectures working on a holistic task such as question answering based on images need to deal with a large gamut of challenges.
% As we strive for holistic tasks
% and open tasks
% such as question answering based on images, we need to deal with a large gamut of challenges.
In this section, we have distilled a few prominent ones that require a joint reasoning over language and visual inputs. We also argue that holistic architectures can benefit from a common sense knowledge.
Finally, we discuss challenges in data acquisition and show how the task differs from other well known tasks.

\paragraph{Vision and language}
{\it Scalability:}
Vision and language systems ground any internal representation in an external world that serves as a common reference point for machines and humans.
The human conceptualization divides these percepts into different instances, categories as well as spatio-temporal concepts. Architectures that aim at reproducing this space of human concepts need to capture the same diversity and therefore scale up to thousands of concepts. 
% \citep{wsabie,perronnin2012towards,Hoffman14Lsda}.
\\
{\it Concept ambiguity:} As the number of categories grows, the semantic boundaries become more fuzzy, and hence ambiguities and gradual memberships are inherently introduced. 
% \cite{lakoff1990women}.
For instance, difference between 'night stand' and 'cabinet', or 'armchair', 'chair' and 'sofa' can be blurry.
Such ambiguities are challenging in at least two ways. Methods need to distinguish fine-grained differences between these objects when appropriate. 
Objective functions and evaluating metrics need to gradually penalize the methods for their mistakes.
% Teachers and magistrates need to gradually penalize the learners for their mistakes.
%
% Therefore it is reasonable to expect from the holistic architectures to create alternative hypotheses of the external world during inference.
% This also relates to the gradual category membership in human perception as portrayed in the prototype theory \cite{lakoff1990women,rosch1973natural}.
\\
% {\it Attributes: }The human concepts are not  limited to  object categories, but also include attributes such as genders, colors, states (lights can be either on or off).
% Often these concepts cannot be learned on their own, but rather are contextualized by the associated noun. E.g. white in ``white'' elephant is surly different from ``white'' in white snow.\\
{\it Ambiguity in reference resolution:} 
% Reliably answering on questions is challenging even for humans. 
The quality of an answer depends on how ambiguous and latent notions of reference frames and intentions are understood \citep{malinowski14nips}.
% Reliably resolving references is challenging -- sometimes even humans need clarification -- but an important part of the question-answering problem, where the architecture's answer depends on the proper resolution. Ambiguities can be caused by a latent notion of reference frame and intentions \cite{malinowski14nips,golland2010game}.
Depending on the cultural bias and the context, we may use object-centric or observer-centric or world-centric frames of reference \citep{levinson2003space}. Moreover, it is no unified notion what 'with', 'beneath', 'over' mean. 
% It seems at least difficult to symbolically define them in terms of predicates.
% While holistic learning and inference encompassing all the aforementioned aspects has yet to be shown, current research directions show promise \cite{beltagy2013montague,rocktaschellow,lewiscombining} by adapting the symbolic-based approaches  \cite{zettlemoyer2007online,kwiatkowski2010inducing,liang2013learning,berant2014semantic} with vector-based approaches \cite{mikolov2013distributed,socher2013grounded, iyyer2014neural} to represent the meaning.
%A holistic learner could use the former while sensing and latter during the complex reasoning \cite{harnad1990symbol,krishnamurthy2013jointly,malinowski14nips}.

\paragraph{Common sense knowledge}
Interestingly, some questions can be quite reliably answered with  access to common sense knowledge. For instance "Which object on the table is used for cutting?" already narrows down the likely options 
significantly. % and the correct answer is probably
%to
%``knife'' or ``scissors''. 
Such example suggests that question-answering architectures would significantly benefit from common sense knowledge.

% It turns out that some questions can solely be answered with the access to common sense knowledge with high reliability. For instance "Which object on the table is used for cutting?" already narrows the likely options significantly and the correct answer is probably ``knife'' or ``scissors''. Other questions like "Which hand of the teacher is on her chin?" require the mixture of the vision and language. To understand the question, a holistic learner needs to first detect a person, figure out that the person may be a teacher, understand a gender of the person, detect her chin, understand 'left' and 'right' side, and finally relates 'her' with the 'teacher'.

%However, different parts of common sense knowledge can be used with different modality.
An 'object for cutting' is not directly visual but about the affordance of the object and therefore a challenging concept to acquire from images only. 
On the other hand, co-occurrences in visual data can represent a kind of  visual common sense knowledge of very mundane facts or probabilistic relations that are rarely found in common sense knowledge bases.

\paragraph{Annotations}
We argue that despite the aforementioned challenges, ``question answering about images'' has unique advantages over other tasks in terms of data acquisition and task evaluation.
In contrast to grounding 
% (understood here as a logical representation of a physical object)
\citep{krishnamurthy2013jointly},
annotating images with question and answer pairs does not require a detailed annotations of whole scenes in terms of predicates representing objects and their relations. The task is also agnostic to the internal representation of a method. In contrast to language generation \citep{karpathy2014generate, donahue2014long}, the output space of a question answering task is more restricted and hence evaluation of different architectures on the task is easier to formulate. In contrast to typical computer vision tasks like object detection \citep{Everingham10}, architectures are judged solely on right answers, not an internal representation. 
%In principle the task doesn't penalize for using different geometrical shapes of detections (squares, rectangles, circles, etc.). 
In contrast to the traditional Turing Test \citep{turing1950computing}, ``answering questions about images'' is less prone to over-interpretations via associating a meaning to machine answers by the human interrogator. %requiring a more thorough understanding of vision and language. 
Hence, a method can be forced to answer to the point rather than ``cheating'' by giving generic answers or output that is open to interpretations.
 % and hence benchmark results better reflects progress of the field.

%
% \input{concrete_challenges}
% 
\vspace{-0.3cm}
\section{Evaluation of architectures}
\label{section:benchmarks}
% To measure the progress, we seek a metric that compares different architectures on an end-to-end task.

Measuring progress on holistic tasks require identifying its goals. For instance a suitable metric for ``question answering about images'' should evaluate architectures based on produced answers but not on intermediate results such as detections or logical forms.
For a Visual Turing Challenge, we seek a metric that satisfies several properties. The most important are:
 % A suitable metric on a question answering task should evaluate architectures based on produced answers but not on intermediate results such as detections or logical forms.
% 
% Despite the challenges to evaluate different architectures on holistic tasks, the question answering based on visual input task provides a common ground as no intermediate results such as detections or proper logical representation are evaluated. Although such task evaluates architectures only on answers that are easily crowdsourced, allowing free natural language answers leads to another AI problem of judging the correctness of answers.
% Due to ambiguities in human interpretations, quantifying performance becomes challenging. Hence we seek a metric that deals with:
% 
% Complex holistic tasks require a carefully crafted benchmarks: performance metrics to measure progress on the task, and different experimental scenarios. We discuss them in this section.
%
% \paragraph{Challenges of defining a performance metric}
% Together with increasing complexity and openness of the task, quantifying performance of the holistic architectures becomes challenging due to several issues:
\\
{\it Automation:}
Evaluating answers on such complex tasks as answering on questions requires a quite deep understanding of natural language, involved concepts and hidden intentions of the questioner. The ideal but impractical metric would be to manually judge every single answer of every architecture individually. Therefore, we are seeking an automatic approximation so that we can evaluate different holistic architectures at scale. \cite{malinowski14nips} proposed to restrict the answer space in order to achieve this goal, while leaving the questions unconstraint.
\\
{\it Social consensus:}
The complex tasks that we are interested in are inherently ambiguous. The ambiguities stem from many factors such as cultural bias, different frame of reference and fined grained categorization. This implies that multiple interpretations of a question are possible. 
To deal with different interpretations of words, \cite{malinowski14nips} define a WUPS scores using lexical databases \citep{miller1995wordnet} with Wu-Palmer similarity \citep{wu1994verbs}. To deal with different interpretations of a  question, \cite{malinowski14visualturing} suggest that the quality of answers should be measured according to the social consensus where the answers are evaluated against multiple ground-truths. Interestingly, such metric also naturally quantifies social agreement of the answer, and serve as a practical approximation of tedious manual evaluation.

\noindent{\bf Experimental scenarios}
In many cases, success on challenging learning problems has been accelerated by use of external data in the training.
% , e.g. in object detection \cite{girshick2014rcnn}.
We believe that a Visual Turing challenge should consists of a sub-task with a prohibited use of auxiliary data to understand how the holistic learners generalize from limited and challenging data in a more established setup. On the other hand, we should not limit ourselves to such artificial restrictions in building the next generation of the holistic learners. Therefore open sub-tasks with a permissible use of additional sources in the training have to be stated, including: additional vision and language resources, synthetic data and curated data. 
% Finally, \cite{malinowski14nips} suggests small and large scale experiments. Former uses a subset of $25$ test images with $37$ object categories in the answers. The large scale experiment challenges scalability of the architectures and uses  $654$ test images with $894$ object categories in the answers.
% 
\vspace{-0.3cm}
\section{Summary}
The goal of this contribution is to sparkle discussions about challenges and benchmarking architectures on holistic tasks.
% We also point out advantages of question answering based on visual input task over other tasks such as traditional Turing Test, grounding, and object detection.
We also argue that ``question answering about images'' is a holistic task that offers multiple advantages over related tasks. For example, it is likely to be less prone to ``cheating'' by over-interpretations than a traditional Turing Test, the annotation process is tractable by crowdsourcing question and answer pairs, and the task does not artificially force any internal representation on the methods.
 % , where the annotation process is tractable, and point out advantages of such task over traditional Turing Test, grounding, and object detection.
Our most recent efforts and results on establishing a Visual Turing Test can be found on our website: 
 % \href{http://www.kcp.krakow.pl}{\nolinkurl{www.kcp.krakow.pl}}
% \url{http://www.d2.mpi-inf.mpg.de/visual-turing-challenge}
\href{http://www.d2.mpi-inf.mpg.de/visual-turing-challenge}{\nolinkurl{www.d2.mpi-inf.mpg.de/visual-turing-challenge}}.

% The goal of this contribution is to sparkle discussions about benchmarking holistic architectures on complex and more open tasks. 
% We identify particular challenges that holistic tasks should exhibit and exemplify how they are manifested in a recent question answering challenge \cite{malinowski14nips}.
% To judge competing architectures and measure the progress on the task, we suggest several directions to further improve existing metrics, and discuss different experimental scenarios.
%The most recent benchmarks together with the challenge can be found in our website \footnote{\url{http://www.d2.mpi-inf.mpg.de/visual-turing-challenge}}.

% The metrics embrace many human answers in the judgement.
% Finally, we also discuss different experimental scenarios.

% 
\vspace{-0.3cm}
\bibliographystyle{apalike}
\small
\bibliography{egbib}

\begin{thebibliography}{}

\bibitem[Donahue et~al., 2014]{donahue2014long}
Donahue, J., Hendricks, L.~A., Guadarrama, S., Rohrbach, M., Venugopalan, S.,
  Saenko, K., and Darrell, T. (2014).
\newblock Long-term recurrent convolutional networks for visual recognition and
  description.
\newblock {\em arXiv:1411.4389}.

\bibitem[Everingham et~al., 2010]{Everingham10}
Everingham, M., Van~Gool, L., Williams, C. K.~I., Winn, J., and Zisserman, A.
  (2010).
\newblock The pascal visual object classes (voc) challenge.
\newblock {\em International Journal of Computer Vision}, 88(2):303--338.

\bibitem[Karpathy and Fei-Fei, 2014]{karpathy2014generate}
Karpathy, A. and Fei-Fei, L. (2014).
\newblock Deep visual-semantic alignments for generating image descriptions.
\newblock {\em arXiv:1412.2306}.

\bibitem[Karpathy et~al., 2014]{karpathy2014deep}
Karpathy, A., Joulin, A., and Fei-Fei, L. (2014).
\newblock Deep fragment embeddings for bidirectional image sentence mapping.
\newblock In {\em NIPS}.

\bibitem[Krishnamurthy and Kollar, 2013]{krishnamurthy2013jointly}
Krishnamurthy, J. and Kollar, T. (2013).
\newblock Jointly learning to parse and perceive: Connecting natural language
  to the physical world.
\newblock {\em TACL}.

\bibitem[Krizhevsky et~al., 2012]{krizhevsky2012imagenet}
Krizhevsky, A., Sutskever, I., and Hinton, G.~E. (2012).
\newblock Imagenet classification with deep convolutional neural networks.
\newblock In {\em NIPS}.

\bibitem[Levinson, 2003]{levinson2003space}
Levinson, S.~C. (2003).
\newblock {\em Space in language and cognition: Explorations in cognitive
  diversity}, volume~5.
\newblock Cambridge University Press.

\bibitem[Liang et~al., 2013]{liang2013learning}
Liang, P., Jordan, M.~I., and Klein, D. (2013).
\newblock Learning dependency-based compositional semantics.
\newblock {\em Computational Linguistics}.

\bibitem[Malinowski and Fritz, 2014a]{malinowski14nips}
Malinowski, M. and Fritz, M. (2014a).
\newblock A multi-world approach to question answering about real-world scenes
  based on uncertain input.
\newblock In {\em NIPS}.

\bibitem[Malinowski and Fritz, 2014b]{malinowski14poolingretrieval}
Malinowski, M. and Fritz, M. (2014b).
\newblock A pooling approach to modelling spatial relations for image retrieval
  and annotation.
\newblock {\em arXiv:1411.5190}.

\bibitem[Malinowski and Fritz, 2014c]{malinowski14visualturing}
Malinowski, M. and Fritz, M. (2014c).
\newblock Towards a visual turing challenge.
\newblock In {\em Learning Semantics (NIPS workshop)}.

\bibitem[Miller, 1995]{miller1995wordnet}
Miller, G.~A. (1995).
\newblock Wordnet: a lexical database for english.
\newblock {\em CACM}.

\bibitem[Turing, 1950]{turing1950computing}
Turing, A.~M. (1950).
\newblock Computing machinery and intelligence.
\newblock {\em Mind}, pages 433--460.

\bibitem[Wu and Palmer, 1994]{wu1994verbs}
Wu, Z. and Palmer, M. (1994).
\newblock Verbs semantics and lexical selection.
\newblock In {\em ACL}.

\end{thebibliography}

\end{document}